\documentclass[transmag]{IEEEtran}
\usepackage{latexsym}
\usepackage{graphicx}
\usepackage{epsf}
\usepackage{caption}
\usepackage{graphicx}
\usepackage{amsmath}
\usepackage{amssymb}
\usepackage{epsfig,latexsym,amsmath,epsf,amssymb,amsfonts}
\usepackage{algorithm, algorithmic}
\usepackage{placeins}
\usepackage{array,color}
\usepackage{cite}
\usepackage{textcomp}
\usepackage{makecell}
\usepackage{soul}
\usepackage{hyperref}
\def\BibTeX{{\rm B\kern-.05em{\sc i\kern-.025em b}\kern-.08em T\kern-.1667em\lower.7ex\hbox{E}\kern-.125emX}}
\begin{document}

\title{Deep Reinforcement Learning for Distributed and Uncoordinated Cognitive Radios Resource Allocation}

\author{Ankita Tondwalkar and Andres Kwasinski,
\IEEEmembership{Senior Member, IEEE}
\thanks{Ankita Tondwalkar is with the Kate Gleason College Of Engineering, Rochester Institute of Technology, Rochester, NY, 14623, USA. (e-mail: at3235@g.rit.edu).}
\thanks{Andres Kwasisnski is with the Department Of Computer Engineering,  Rochester Institute of Technology, Rochester, NY, 14623, USA. (e-mail: axkeec@g.rit.edu).}
}
\IEEEtitleabstractindextext{
\begin{abstract} This paper presents a novel deep reinforcement learning-based resource allocation technique for the multi-agent environment presented by a cognitive radio network where the interactions of the agents during learning may lead to a non-stationary environment. The resource allocation technique presented in this work is distributed, not requiring coordination with other agents. It is shown by considering aspects specific to deep reinforcement learning that the presented algorithm converges in an arbitrarily long time to equilibrium policies in a non-stationary multi-agent environment that results from the uncoordinated dynamic interaction between radios through the shared wireless environment. Simulation results show that the presented technique achieves a faster learning performance compared to an equivalent table-based Q-learning algorithm and is able to find the optimal policy in 99\% of cases for a sufficiently long learning time. In addition, simulations show that our DQL approach requires less than half the number of learning steps to achieve the same performance as an equivalent table-based implementation. Moreover, it is shown that the use of a standard single-agent deep reinforcement learning approach may not achieve convergence when used in an uncoordinated interacting multi-radio scenario.\end{abstract}

}

\maketitle

\section{INTRODUCTION}

\IEEEPARstart To keep up with the ever increasing performance demands from wireless applications, wireless networks necessitate to operate following an efficient use of the available radio spectrum. Dynamic spectrum access (DSA) and sharing,\cite{ref1}, \cite{ref2}, \cite{ref3},\cite{ref4}, \cite{ref5},\cite{ref6},\cite{ref7}, plays a crucial role in improving the utilization of the radio spectrum, as it departs from the traditional approach of static radio spectrum band allocation which has led to spectrum under-utilization. Cognitive radios (CRs) are considered as the answer to realize this more efficient and effective use of the radio spectrum because of their ability to autonomously gain awareness of the wireless network environment and learn to adapt to changing conditions. Due to its model-free characteristic, reinforcement learning (RL) is a machine learning approach to resource allocation that naturally aligns with the CR paradigm \cite{ref8}, \cite{ref9}. However, the application of RL in general wireless networking present a key challenge. In a general scenario with distributed and uncoordinated transmissions (e.g. an ad-hoc network), the transmission from one CR affects the environment perceived by other CRs (appearing as interference) resulting in a multi-agent non-stationary environment. The multi-agent interaction with the non-stationary environment and the entanglement of agents due to their transmissions may not necessarily lead to the convergence of RL to the best policy, as it is guaranteed after arbitrarily long learning time in the case of single agent Q-learning \cite{ref10}. In fact, it may be the case that it is not possible to define an optimal policy for all the multiple agents and, instead, it is necessary to consider convergence to some form of equilibrium state (e.g. Nash equilibrium). 

In this work, we consider the challenging case of a multi-agent non-stationary environment where nodes in a CR network (CRN) share the spectrum with a primary network (PN) by operating in an uncoordinated distributed fashion following an underlay DSA scheme,\cite{ref7, ref11, ref12}. Specifically, the goal of this work is to develop a distributed and uncoordinated resource allocation RL mechanism for entangled CRs, this is, where a change of transmission parameters of one radio affects the operating environment and performance of the others. To provide a practical context to the study, we choose transmit power control as the resource to manage. As such, this work belongs to a class of notoriously challenging distributed multi-agent control problems known as Weakly Acyclic Stochastic Dynamic Games. Until recently, for this class of problems there was no known RL algorithm with guaranteed convergence to the equilibrium policy. However, \cite{ref13} recently presented a modified general table-based Q-learning algorithm (a very common form of RL) for which convergence in an asymptotic infinite learning time was proved.

At the same time, the research area of RL has seen notable advances over the recent years. The work \cite{ref14} constitutes a major milestone by introducing deep Q-networks (DQNs), a new approach to Q-learning with better learning performance than conventional table-based Q-learning. This approach is based on the use of deep neural networks to approximate the Q action-value function. This advance spurred research into the application of single-agent DQNs for a variety of applications, including wireless communications. In \cite{ref15}, a single-agent power control Deep Q-learning (DQL) technique was introduced for an underlay CR system consisting of a single primary and secondary link. There is an exchange of information between the two networks, which helps the secondary user to adjust its transmit power levels. The work \cite{ref16} applied DQL to dynamic resource allocation in cloud Radio Access Networks, \cite{ref17} studied the application of DQL to control interference alignment, and \cite{ref18} used DQL as a trainable function approximator for arbitrary resource allocation algorithms. The work in \cite{ref19} presents a power allocation technique for a cellular network using DQL based on training at a central node, and \cite{ref20} also presents a DQL technique with centralized training based on the experiences gathered from all agents. Unlike \cite{ref19} or \cite{ref20}, our proposed solution does not rely on the centralized training for convergence to optimal policy. Our RL methodology is distributed, resulting in uncoordinated and dynamic interaction between the cognitive radios. The field of work researching multi-agent DQL for wireless communications is rapidly growing \cite{ref3, ref21, ref22, ref23, ref24, ref25} but still sparse and has not majorly focused on scenarios as the one considered herein with a non-stationary environment because they usually allow agent coordination or learning at a central node.

Our main contribution in this paper is to present a multi-agent DQL technique for distributed resource allocation in a CRN with no requirement for agent's coordination. First we show that the the DQL learns faster than an equivalent conventional table-based approach. Second, we demonstrate that the algorithm converges to the optimal policy in the limit of infinite running time. The convergence in case of our RL algorithm happens almost surely in stochastic environments. We also explain how the variance of the estimation error of Q-values that is introduced due to the non-stationary environment can be reduced to an arbitrarily small value. Moreover, we show that our distributed and uncoordinated solution converges to an optimal solution in scenarios where it is possible to define such a solution. Some of the contributions of our technique are as follows:
 \begin{itemize}
    \item Our technique is the first to present a multi-agent DQL technique for uncoordinated and distributed CR resource allocation with proven convergence (with probability one) to deterministic equilibrium policy. The non-stationarity arising from the lack of agent coordination makes it difficult to define an optimum, hence we have modified our network setup to follow properties in terms of optimal solution for the equilibrium state.
    \item The key challenge addressed by our technique is accounting for the presence of multiple active CR agents leading to a non-stationary environment due to the interaction of the learners through the shared wireless environment.
    \item We analytically show that our proposed DQL technique can achieve convergence when learning time is allowed for arbitrarily long learning time (as is customary in RL). This result is confirmed through simulation results that show that our technique converges to equilibrium in nearly 99\% of cases when learning is allowed to progress for a sufficient long time.
    \item In addition, simulation results will show that our proposed technique is able to learn notably faster than an equivalent table-based implementation, the former requiring less than half the number of learning steps to achieve the same performance as the later.
    \item Moreover, as another key contribution, we present a case that shows that the application of standard single-agent DQL in uncoordinated distributed CR resource allocation may not reach convergence (to any policy, optimal or not) due to the large noise present in Q-value estimation from the non-stationary environment.
 \end{itemize} 
 
The remainder of this paper is organized as follows. We present the system model in Section \ref{syssetupandprob}. In Section \ref{clcr} we provide an overview of deep-Q learning and then discuss in detail the DQN-based distributed and uncoordinated multi-agent resource allocation algorithm. The simulation results and analysis is given in Section \ref{simulation}, followed by conclusions in Section \ref{conclsec}.

%\vspace*{-1mm}
\section{System Setup and Problem Formulation}\label{syssetupandprob} %\vspace*{-1mm}

\begin{figure}[tb]
	\centering
	\includegraphics[width=0.5\textwidth]{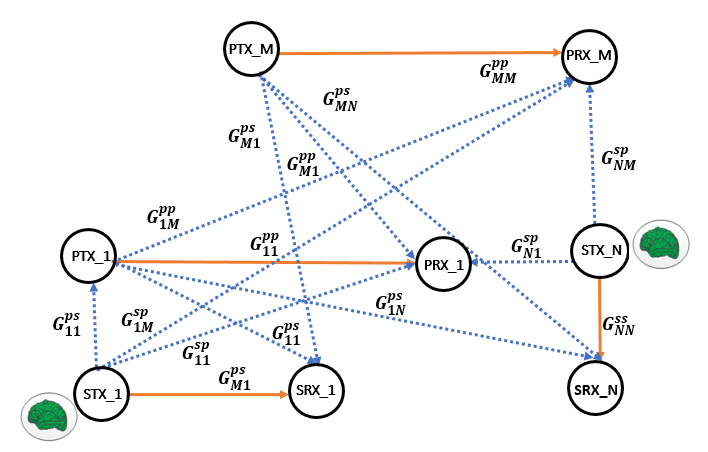}\vspace*{-1mm}
	\caption{Network model for Primary and Secondary network.}
	\captionsetup{aboveskip=0.5pt,font=it}
	\label{Network_model_PN_SN}\vspace*{-1mm}
\end{figure}

We consider a system as illustrated in Fig. \ref{Network_model_PN_SN}, comprised of two networks that operate in a common geographical area and share the same radio spectrum band following an underlay DSA mechanism. In the system, a primary network (PN) with $M$ access points (APs) is incumbent to the spectrum band in use, and a secondary network (SN) is formed by $N$ CRs which are assumed to operate in a fully autonomous manner, i.e. there is no coordination between the CR nodes during resource allocation and no exchange of information between the two networks.
Because of the underlay DSA operation, the CRs in the SN are limited in their transmit power so that the interference they create on the PN does not exceed an established limit. The interference from a CR reduces the Signal-to-Interference-plus-Noise ratio (SINR) and consequently the throughput at PN links. For the system model as shown in Fig. \ref{Network_model_PN_SN}, the SINR ($ \gamma $) for the $i$th PN link, is calculated as,%\vspace*{-0.08in}
\begin{eqnarray}
\gamma^{(p)}_i = \frac{G_{ii}^{(pp)}P^{(p)}_i}{\displaystyle{\sum_{\substack{ j\neq i}}^M}  G_{ji}^{(pp)}P^{(p)}_j + \displaystyle{\sum_{j=1}^N G_{ij}^{(ps)}} P^{(s)}_j + \sigma^2}, \label{eq2_SINRpn} %\\ [-0.2in] \nonumber
\end{eqnarray}
\noindent where $P^{(p)}_i$ is the transmit power at the $i$th active primary link $(i= 1,2,.. M) $,  $P^{(s)}_i$ is the transmit power at the $i$th SN link, $G_{ij}^{(pp)}$  is the channel gain between the $i$th PN AP and the $j$th PN receiver, and  $\sigma^2$ is the background additive white Gaussian noise (AWGN) power. As is customary in today's high performing wireless systems, we assume that the PN employs an Adaptive Modulation and Coding (AMC) technique \cite{ref26} \cite{ref27} that adapts the modulation scheme and the channel coding rate (together known as AMC mode) based on the SINR of transmission link.

The basic premise for the CRs transmission is to achieve as large as possible of an SINR so as to reach the largest possible throughput. The SINR for the $i$th SN link is calculated as,%\vspace*{-0.08in}
\begin{eqnarray}
\gamma^{(s)}_i = \frac{G_{ii}^{(ss)}P^{(s)}_i}{\displaystyle{\sum_{\substack{ j\neq i}}^N}  G_{ij}^{(ss)}P^{(s)}_j + \displaystyle{\sum_{j=1}^M G_{ij}^{(sp)}} P^{(p)}_j + \sigma^2}, \label{eq2_SINRsn} 
\end{eqnarray}
\noindent where $P^{(s)}_i$ is the transmit power at the $i$th active secondary link $(i= 1,..N) $. $P^{(s)}_i$ is the transmit power at the $i$th SN link, $G_{ij}^{(ss)}$  is the channel gain between the $i$th SN cognitive user and the $j$th SN receiver, and $G_{ij}^{(sp)}$ is the channel gain between the $i$th SN cognitive user and a $j$th PN receiver. 

At the same time, the transmit power of CRs is limited by the operation following underlay DSA. Owing to the one-to-one relation between throughput and SINR, the CRs assess their transmission effect on the PN by estimating the relative change in the throughput at its nearest PN link following the procedure explained in \cite{ref28}. Consequently, the CRs implement underlay DSA following a principle equivalent to a limit on interference at the PN where now the limit is set on the relative throughput change at their nearest PN link (understood hereafter to be the one that is received with largest power). Additionally, the operation of the CRs is under the requirement that they cannot access the PN's feedback channel, as this would be a violation of the conditions for uncoordinated interaction between the SU transmit power  two networks. Nodes in both networks perform resource allocation in order to optimize their transmission. 

Therefore, the problem to be studied in this work is the distributed and uncoordinated resource allocation for the cognitive agents in the SN. For the purpose of this work, namely the development of an algorithm where each CR independently learns its best resource allocation decision for transmission without previous knowledge of the wireless environment, we focus on the allocation of transmit power only (because the nature of the problem under study does not change if considering additional transmission parameters). Specifically, transmit power allocation at each CR seeks to achieve the largest possible throughput without exceeding the limit on relative throughput change on their nearest PN link. This problem under study is characterized by CRs acting as learning agents within a framework of uncoordinated multi-agent interactions, where the action of one agent may affect the state of the wireless environment for other agents (transmission of one CR can be a source of interference to the rest of the SN and PN). The absence of coordination between the two networks or even within the SN adds complexity to the already challenging power allocation problem for the cognitive agents. Indeed, the multi-agent interaction results in a non-stationary environment that affects the CRs learning by presenting noisy feedback information on the results of actions (transmit powers) potentially causing non-convergent learning behaviour for the agents. As will be seen, our proposed technique effectively addresses these challenges ensuring convergence to an optimum (when such optimum can be defined).

\section{DQN-Based Distributed and Uncoordinated Multi-Agent Resource Allocation}\label{clcr}

%\subsection{Overview of deep Q-Learning}

The uncoordinated agents in the SN seek to find an allowable transmit power such that the interference constraint on the PN is successfully met. Since we assume that there is no prior knowledge about the radio environment and that there is no exchange of information between the PN and the SN or between the CRs, the solution to the problem at hand needs to enable the CRs to operate as agents able to autonomously learn and adapt to the dynamic environment conditions. The requirement for autonomous learning of the environment conditions can be conceptualized as the need for a model-free approach. Autonomous learning to adapt behavior to the environment can be realized following an Observe-Orient-Decide-Act framework canonical to CRs \cite{ref29}. These considerations lead us to approach the solution to the problem at hand using model-free RL.

\begin{figure}[tb]
	\centering
	\includegraphics[width=0.5\textwidth]{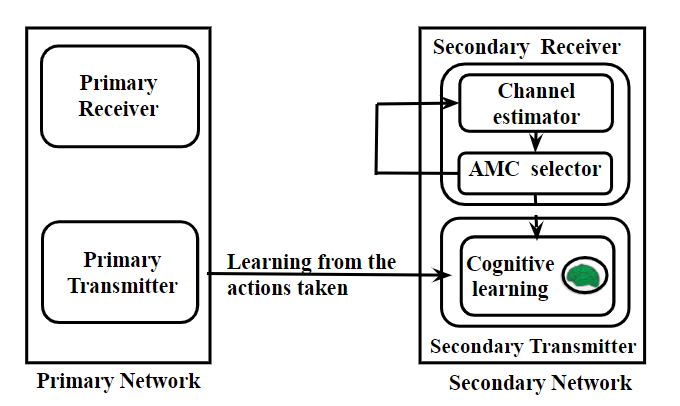}\vspace*{-0.9mm}
	\caption{Network model for Primary and Secondary network.}
	\label{Networkmodel2}\vspace*{-1mm}
\end{figure}

The RL technique enables the agent to observe and learn from its surrounding and helps build a model that is flexible and easily adapts to the  wireless environment dynamics as shown in Fig. \ref{Networkmodel2}. The non-stationarity of the environment as a result of multi-agent interaction as in our case, causes the wireless conditions to vary dynamically. Therefore, it is challenging in practice, to develop an accurate wireless model that adapts to the surrounding changes by estimating its parameters. Model-free RL does not need a prior knowledge about the network dynamics and therefore can be used as a black box modelling approach where the CRs can implicitly learn about the wireless network environment without the need of any predefined model.

Therefore, we will solve the problem described in Section \ref{syssetupandprob} using a RL approach. A RL model is characterized through the definition of three elements: an action space $\mathcal{A}= \{a_0,a_1,\dots,a_m\}$, an state space $\mathcal{S}=\{S_0,S_1,\dots,S_{n-1}\}$, and a reward function $\mathcal{R}$. We assume that the different condition that the environment could be in can be abstracted as a set of states. The state space is the set of all possible states  ${s_t} \in \mathcal{S}$ that the environment can be in. The dynamics of the transition between environment states is unknown for the case of the distributed and uncoordinated multi-agent scenario at hand but it will be influenced by the actions taken by the agents. Through RL, the agents follow a process of trial and error taking actions that progressively enables the learning of what actions are more appropriate for each state. At time step $t$, an agent learning with RL executes an action $a_t \in \mathcal{A} $ and, subsequently, both receives a reward $r_{t+1}$ associated with the result of its action on the environment and observes the next state $s_{t+1}$ to which the environment has transitioned to. In RL, the main goal of an agent is to learn the best policy $\pi^*$ that maximizes the expected cumulative discounted reward. A policy $\pi^*$ is a mapping from the state space to the action space.
\begin{figure}[!tbhp]
	\centering
	\includegraphics[width=0.5\textwidth]{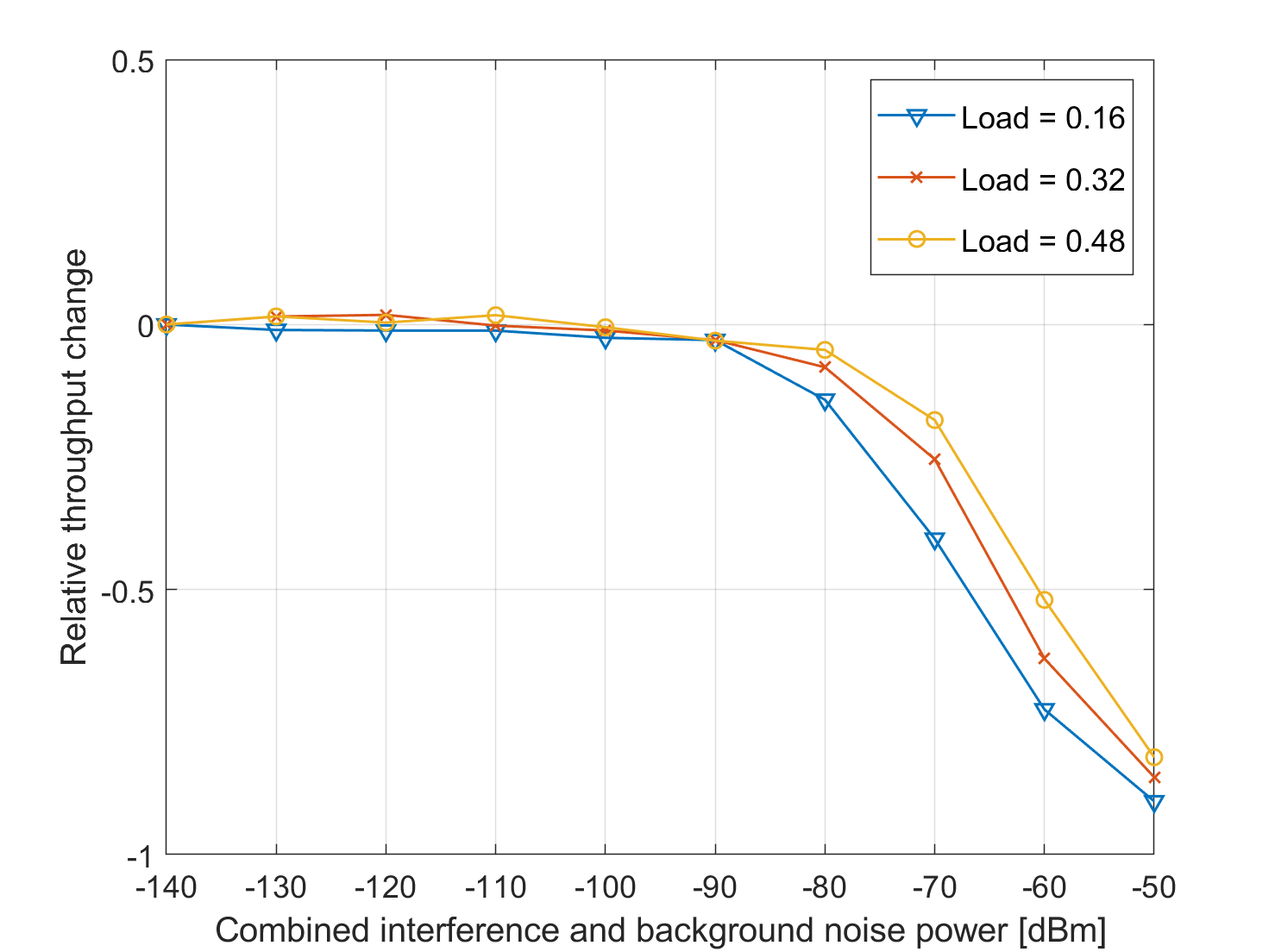}\vspace*{-0.1mm}
	\caption{Relative throughput change vs background noise power in primary network. \cite{ref28}}\vspace*{-0.09mm}
	\label{RelThroughputN}%\vspace*{-0.8mm}
\end{figure}

To apply RL as the solution of our problem we define the action space, state space, and reward function as follows:

\emph{\bf{Action space:}} We define the action space as the set $\mathcal{A}=\{0,P_0,P_1,\dots,P_L\}$ of possible transmit powers (where '0' indicates no transmission), without loss of generality (owing to the uncoordinated distributed operation of the CRs), we assume that all CRs have the same action space. In our uncoordinated and distributed setup, each agent chooses an action independently of the other agents. 

\emph{\bf{State space: }} We define the state space as consisting of two states, $S_0$ and $S_1$, where state $S_0$ corresponds to the conditions when the interference limit for underlay DSA is met, and state $S_1$ corresponds to the case when the condition is not met. As previously indicated, the underlay DSA limit is characterized based on the effect that the SN interference has on the PN through the relative change of throughput. As discussed in \cite{ref28}, a network that transmits using AMC, like the PN in our setup, can absorb up to a limit the increase in the combined background noise and interference without experiencing a significant decrease in average throughput.
This property of AMC is illustrated in Fig. \ref{RelThroughputN}, which shows the relative average throughput change as a function of background noise power for a network transmitting with AMC (the configuration is the same described for the PN later in Sect. \ref{simulation}). Since the underlying principle of our uncoordinated multi-agent setting, is that there is no exchange of any information between the two networks, for the PN the interference from SN is indistinguishable from an increase in the background noise power. However, since the background noise power is usually constant, Fig. \ref{RelThroughputN} effectively is showing the PN relative average throughput change as a function of the increase in the interference from the SN. Moreover, it was shown in  \cite{ref28} that this relative average throughput change in the PN, hereafter denoted as $T_{\%}$, can be expressed as a function of the interference from the SN through the approximate expression,
\begin{eqnarray}
T_{\%} =  -1/\xi * \log_{2}\left(1 + \Gamma * 10^{I/10}\right), 
\end{eqnarray}
\noindent where $\xi$ is the PN spectral efficiency in absence of any SN transmission, $\Gamma$ is the signal-to-noise ratio (SNR) gap accounting for the the use of practical coding and transmission, and $I$ is the interference from the SN received at the PN. Following the operation described in \cite{ref28}, each CR manages its effect on the PN by assessing the relative throughput change at the PN link that is received with highest power (usually the nearest link but not necessarily always the case due to random fading phenomena). Denoting as $T_{\%}^{(i)}$ the PN link relative throughput change assessed by CR $i$ and $\epsilon$ the limit on the relative throughput change, the state space is then defined as,
\begin{eqnarray}\label{eq_state_def}
s_t=\left\{\begin{array} {ll}
S_0, & \textrm{if } |T_{\%}^{(i)}| \leq \epsilon,\\
S_1, & \textrm{else }.
\end{array} \right. 
\end{eqnarray}

\emph{\bf{Reward: }}The reward function reflects each CR goal to maximize their throughput and is defined as a function of the state and the action taken to transition to a new state as,
\begin{eqnarray}\label{eq_Reward}
R^{(i)}_t(a_t, s_t)\!=\!\left\{\begin{array} {ll}
10^{T_i}, & \textrm{if environment state is } S_0,\\%[-.03in]
0, & \textrm{if environment state is } S_1,
\end{array} \right. 
\end{eqnarray}
\noindent where $T_i$ is the throughput on the $i$th SN link. This reward function was designed to augment the difference between rewards for actions that correspond to low SINRs. 
Reinforcement learning algorithms are intended to find the best policy $\pi^*$ that maximizes the cumulative expectation of discounted rewards over time, which is represented as the state-value function, 
\begin{eqnarray}\label{Expected_sum_of_discounted_rewards}
V_\pi(s)=\sum_{t=1}^{\infty} \gamma^{t}E[R_t|\pi,(s_0=s)] %\\ [-.2in] \nonumber
\end{eqnarray}
There exists different RL algorithms both for the model-based as well as the model-free cases. In the model-free cases, the state-value function needs to be estimated because the probability distribution underlying the expectation in \eqref{Expected_sum_of_discounted_rewards} is unknown. In \cite{ref30}, Watkins introduced the now popular Q-learning algorithm to do estimate the action value function, i.e. the Q-values $Q(s_t,a_t)$, which is the state-value function resulting from taking at time $t$ an action $a_t$ in state $s_t$, followed in the remaining time by the sequence of actions that maximizes the state-value function.

A traditional Q-learning algorithm, which we will call the ``table-based'' Q-learning algorithm, organizes the Q-values into a states $\times$ actions table and estimates the Q-values through the time-difference equation:
 \begin{eqnarray}
 \label{TD}
Q(s_t,a_t) \leftarrow  Q(s_t,a_t)+\alpha[R_{t+1}+ \gamma\max_{{a'} }Q(s'_{t+1}, a')
\nonumber\\[-.023in]
-Q(s_t,a_t)]
\end{eqnarray}
where $s'_{t+1}$ is the next state, and $a'$ is the action to be taken in the next state, $\alpha\in [ 0, 1 ]$ is the learning rate, and $\gamma \in [0,1]$ is the discount factor. The table-based Q-learning algorithm presents the shortcoming that it does not scale up well with increasing sizes of the state or action spaces, resulting in very large number of learning steps necessary to accurately estimate the Q-values.

An alternative approach to the table-based Q-learning is to estimate the bi-variate function $Q(s_t,a_t)$ through a regression approach. The implementation of the regression through an artificial neural network model (called a deep Q-network, DQN) with parameters vector $\theta$ doing $Q (s,a;\theta) \approx Q (s,a)$ is what has come to be known as a deep Q-learning (DQL), \cite{ref14}. Moreover, in order to damp and reduce the likelihood of oscillations in the policy during learning, in DQN each agent utilizes two separate neural networks as Q values approximators: one as action-value function approximator $Q_i(x,{a};\theta_i)$ and another as target action-value function approximator $\hat{Q_i}(x,{a};\theta^-_i)$, where $\theta_i$ and $\theta^-_i$ denote the parameters (weights) for each of the neural networks. At each learning step, the parameters $\theta_i$ of each agent's action-value function are updated from a mini-batch of training samples each consisting of tuples $<$\emph{current state, next state, action taken, target Q-value}$>$ using a gradient descent back propagation algorithm with the error function,
\begin{eqnarray}
\label{grad_dec}
L(\theta_i)\!\!=\!\frac{1}{2} [r_i(x,a)\!+\!\gamma \underset{\hat{a}\in A}{\mathrm{\textstyle\max}}\: \hat{Q}_i(\hat{x},\hat{a};\theta^-_i) \!-\! Q_i(x,{a};\theta_i)]^2,\!\! 
\end{eqnarray}
where $r_i(x,a)$ is the observed reward after taking action $a$ while in state $x$. Only every $c$ learning steps, the parameters $\theta^-_i$ of the target action-value function are updated by copying the updated parameters $\theta_i$ of the action-value function. 

\begin{figure}[tb]
	\centering
	\includegraphics[width=0.5\textwidth]{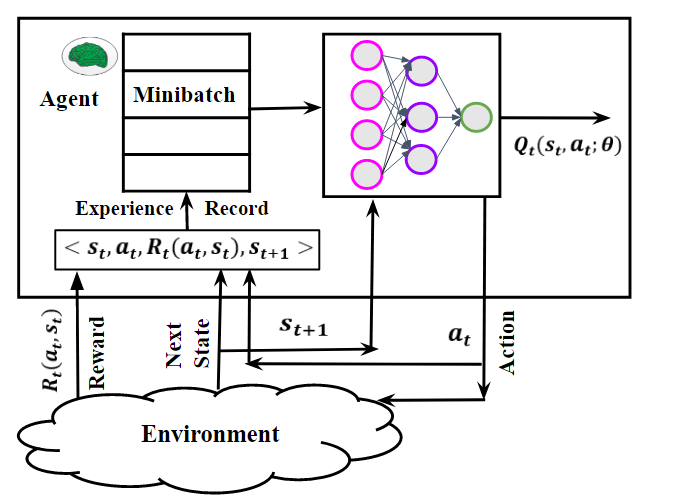}\vspace*{-0.01mm}
	\caption{System model for Primary and Secondary network.}
	\label{SystemModel}
\end{figure}

Compared to table-based Q-learning, the approach followed with DQL of approximating the action-value function through a regression presents the advantage of faster learning when the action and/or state spaces are large. It is because of this better performance that we adopt DQL as the basis for the resource allocation technique. The application of DQL as explained above, which we will identify as ``standard DQL'', has shown significant success in many applications, especially for a single agent setting. The standard DQL technique shares with single agent Q-learning the property of a guaranteed convergence to the optimal policy as time tends to infinity. However, the problem at hand is one of multiple agents where actions of any one of them may affect the environment of other agents (the transmission of one CR translates into interference to the rest of the SN and the PN), which, combined with a lack of agent coordination, may result in a non-stationary environment under which the standard DQL may not necessarily yield an optimal result.

In terms of non-stationary elements that affect learning, standard DQL already incorporates a mechanism to address non-stationarity of the action-value  function during learning, which would affect the second term in the target value of the loss function (i.e. $\gamma \max_{\hat{a}} Q_i(\hat{x},\hat{a};\theta_i)$) conventionally used in the DQN weights learning through back propagation. This issue is addressed through the introduction of the previously described target action-value approximator, which is updated every $c$ learning steps, leading to the loss function shown in \eqref{grad_dec}. Unfortunately this mechanism does not tackle the non-stationarity of the environment arising out of the multi-agent interaction. This non-stationarity of the environment affects now the first term of the target value in the loss function \eqref{grad_dec} by resulting in ``noisy'' and non-stationary observations of the rewards $r_i(x,a)$. As a result of the noisy rewards, the learning of the DQN may not converge. To address this unstable environment issue in table-based Q-learning, \cite{ref13} introduced the idea that learning occurs over a succession of multiple ``exploration phases'' where in each of these phases the agents take the action determined by the current policy most of the time and only occasionally (as determined by a random draw based on an ``experimentation probability'') it explores a different action. This results in agents experiencing near-stationary environments (with stationarity broken only occasionally by the exploration of non-current policy actions by other agents), which allows the agents to accurately estimate their Q-values. Moreover, in addition to the mechanism of exploration phases, it is necessary to also implement a mechanism that does not require coordination between agents for the gradual learning of the best policy. In \cite{ref13}, the technique proposed for this purpose is a mechanism called "Best Reply Process with Inertia". In this mechanism, after each exploration phase, a policy is selected from a candidate set formed from those actions associated with the Q-values that are within a tolerance range of the largest Q-value. Inertia in the choice is implemented by keeping as policy the current action if it belongs to the candidate set or, otherwise, keeping as policy the current action with an inertia probability $\lambda$ (usually larger than 0.5), and choosing one policy with probability $1-\lambda$ from the candidate set.

Consequently, we propose a novel variation of DQL (Fig. \ref{SystemModel}) that is apt for distributed uncoordinated multi-agent scenarios by incorporating the mechanisms of exploration phases and the selection of a policy with inertia. Algorithm \ref{alg1} shows the proposed DQL technique. In the algorithm, lines \ref{startexp} to \ref{endexp} comprise an exploration phase. Within one exploration phase, line \ref{chooseactalg} shows the choice of action and line \ref{DQNlearn} comprises the action-value function DQN approximator learning. Lines \ref{startinert} and \ref{endinert} show our implementation of the mechanism to choose a policy with inertia which differs from the implementation in \cite{ref13}. In our implementation the current policy is kept with probability $\lambda$ and a new policy is chosen from the set of candidates (which may or may not still include the current policy) with probability $1-\lambda$. Also, the tolerance level used to determine the set of candidate policies $\delta^i$ in line \ref{startinert} is also particular to our solution, because it is not constant as in \cite{ref13} but instead it is designed to adapt to the ``noisy'' estimates of Q-values, by being calculated as a number of times (three times in our implementation) the largest moving standard deviation among Q-values. Also, in our DQL implementation we eliminate the replay memory that is present in standard DQL because it operates counter to the intend of an exploration phase. In addition we eliminated the target action-value DQN, replacing it by an array that is updated in line \ref{updatetavf} every $c$ step with Q-values calculated using the action-value function DQN approximator.

\begin{algorithm}[tb]
 	\caption{Uncoordinated Distributed Multi-agent DQL}\vspace*{-.3mm} \label{alg1}
 	\begin{algorithmic}[1]
 	    \STATE Set parameters \vspace*{-.4mm}
 	    \STATE $\rho\in$ {(0,1)} : {Experimentation probability} \vspace*{-.4mm}
 	    \STATE $\lambda\in$ {(0,1)} : {Inertia} \vspace*{-.4mm}
 	    \STATE $\gamma\in$ {(0,1)} : {Discount factor}
 	    \STATE $L_E\in \{1,2,3,\dots\}$ : {Exploration phase length}
 	    \STATE Initialize policy $\pi_0 \in \Pi$ {(arbitrary)} \vspace*{-.4mm}
        \STATE Sense state $s_0$ \vspace*{-.4mm}
 		\STATE 	Initialization of the neural network for action-value function $Q_i$ with weights $\theta_i$ randomly chosen with uniform distribution between 0 and 1. \vspace*{-.4mm}
 		\FOR { $0\leq k \leq K$}
 		\STATE $t_k=kL_E$
 	    \FOR {Iterate $t= t_k, t_k +1,\cdots, t_k +L_E-1$}\label{startexp}
 	    \STATE ($k$th. exploration phase)\\
 	            $ a_{t} =
        \begin{cases}
       \text{$\pi_k(s_t), \hspace{2mm} w.p.\hspace{2mm} 1-\rho $ }\\[-.05in]
       \text{$ \textrm{any } a \in {A},\hspace{2mm} w.p.\hspace{2mm} \rho/|A| $}
       \end{cases} $ \label{chooseactalg}
 		\STATE Update the state $s^{(i)}_{t+1}$  and the reward $R^{(i)}_t$.
 		\STATE Update adaptive learning rate $\alpha_t=\alpha_{t-1}/\zeta$, where $\zeta$ is a constant;
 		\STATE Update parameters ($\theta$) of action-value function $Q(s_t^{(i)}, a_t^{(i)};\theta_i)$, through mini-batch back-propagation with error function \eqref{grad_dec}  \label{DQNlearn} \vspace*{-.4mm}
 		\STATE Once every $c$ learning steps:\\ $\qquad \hat{Q}(s,a) \leftarrow Q(s,{a};\theta_i)$, $\forall \: s, a$. \label{updatetavf} \vspace*{-.2mm}
        \ENDFOR \label{endexp} \vspace*{-.4mm}
        \STATE
        $\Pi_{k+1}^i = \{\hat{\pi}^i \in\Pi^i:Q_{t_{k+1}}^i(s,\hat{\pi}^i(s))$\\
        $\qquad \qquad \ge\max_{v^i}Q_{t_{k+1}}^i (s, v^i)-\delta^i, \text {for all s}\}$ \label{startinert}
        \STATE $ \pi_{k+1}^i =
        \begin{cases}
       \text{$\pi_k^i, \hspace{4mm} w.p.\hspace{4mm} \lambda $ }\\[-.05in]
       \text{$ \mbox{any } \pi^i \in \Pi_{k+1}^i, w.p. \frac {(1-\lambda)} {|\Pi_{k+1}^i|}$}
       \end{cases} $\label{endinert}
        \ENDFOR \vspace*{-.4mm}
 	\end{algorithmic}
 	\label{1}
 \end{algorithm}

\section{Convergence of Proposed Distributed Uncoordinated Multi-agent DQL Algorithm}\label{sect:convergence}
We study in this Section the convergence properties of our proposed distributed and uncoordinated multi-agent DQL algorithm. In this study there are two main issues that need to be considered. The first of these issues is the convergence of Q-learning in the case of a multi-agent scenario. The second issue is how the DQL approach of approximating the bi-variate function $Q(s_t, a_t)$ is affected by the ``noisy'' and non-stationary observations of the rewards. Note that in the general case of a multi-agent scenario, as the one considered here, the convergence is studied in terms of reaching equilibrium states with properties that relate to large rewards across all agents (where states in this context refer to the results of the learning process, not the environment, and are representing a policy). Specifically, we will consider equilibrium states, which are those terminal states reached by following during learning a sequence of strict best replies. Denoting as $\pi^i$ an agent $i$'s policy and as $\pi^{-i}$ the policy of the rest of the agents, a strict best reply with respect to $(\pi^i,\pi^{-i})$ is the agent $i$'s policy  $\hat{\pi}^i$ that achieves agent $i$'s largest reward given $\pi^{-i}$ for all initial states and results in a strict improvement over $\pi^i$ for at least one initial state. Considering the sequence of strict best replies $\pi_1, \pi_2, \dots, \pi_k, \dots$, an equilibrium state is a policy $\pi_k$ from which a subsequent strict best reply with respect to $\pi_k$ does not exist. In the particular case where it is possible to define an optimal policy, the equilibrium state should be that of the optimal policy. 

It is well known that single-agent Q-learning has guaranteed convergence to an optimal policy when each state is visited, and each action is taken, an infinite number of times during learning, \cite{ref31}. However, the single-agent conditions for which this result holds are different from the case of our proposed DQL algorithm. Our problem of uncoordinated and distributed multi-CR resource allocation is a form of a weakly acyclic stochastic dynamic game, a class of specially challenging RL problems for which there was no known RL algorithm with guaranteed convergence to the optimal policy until Arslan and Y\"uksel presented in \cite{ref13}  a table-based Q-learning algorithm for weakly acyclic stochastic dynamic games with proven convergence with probability one in an asymptotically infinite learning time. Specifically, it was shown in \cite{ref13} that for a multi-agent system where agents have access to the environment state information but do not have access to any information about the other agents (not even their actual presence or not), the Q-values estimated through a table-based Q-learning algorithm that relies on exploration phases and the best reply process with inertia converge to the optimal $Q^*(s_t,a_t)$ with probability one for every state and action, i.e.
\begin{eqnarray}\label{Prob}
P[Q(s_t,a_t)\!\rightarrow\! Q^*(s_t,a_t), \pi^* \in \Pi]\!=\!1 \;\; \forall (s_t,a_t) \in \mathcal{S}\times\mathcal{A},\\ [-.2in] \nonumber 
\end{eqnarray}
through the incremental update \eqref{TD}, for any Q-values initial condition, when the agents visit each state action pair infinitely many times and when the learning rate $\alpha_t$, $t\ge 0$, $\alpha_t \in (0,1)$, over time steps $t$ satisfy 
\begin{eqnarray}
\sum_{n}\alpha_n={\infty}; \hspace{1.5cm} \sum_{n}\alpha_n^2<{\infty}, 
\end{eqnarray}

The convergence of estimates to the optimal Q-values in \eqref{Prob} ensures that the learning process across the multiple agents will end at a deterministic equilibrium policy. At its core, this result follows from the ability to turn the non-stationary environment into a near-stationary one through the control of the experimentation probability $\rho$. The smaller that $\rho$ is chosen the more stationary that the environment appears. Our multi-agent DQL maintains all the elements of the table-based algorithm in \cite{ref13} that are needed for the convergence \eqref{Prob}, thus also inheriting the guarantee of ending at a deterministic equilibrium policy. Nevertheless, in table-based Q-learning the Q-values are estimates calculated using an incremental update approach that is driven by the observation of a reward received when the agent is at some state and takes an action. Deep Q-learning follows a different approach to estimate Q-values where the bi-variate function $Q(s_t, a_t)$ is approximated by gradually minimizing the error \eqref{grad_dec} between the current approximation and the observation of a point of the function $Q(s_t, a_t)$, which depends on the observed reward. Therefore, we need for our DQL algorithm to study the effect that ``noisy'' rewards could have on the approximation of the function $Q(s_t, a_t)$ and how it can be ensured that our DQL is able to estimate the deterministic equilibrium policy $\pi^*$.

We have described the observed rewards as being ``noisy'' to succinctly indicate their nature as a measurement of a random process. Our next task is to model this random process. Recall that the reward has been defined in \eqref{eq_Reward} as a function of the throughput on a SN link. This throughput depends on the SINR on the same SN link as given in \eqref{eq2_SINRsn}. Additionally, the reward depends on the environment state. Therefore, randomness into the observation of the rewards is introduced by the random choice of actions of the other agents which manifests as a change in the interference to the SN link (second term in the denominator of \eqref{eq2_SINRsn}) or a change in the environment state (which we will call hereafter as an environment ``change of phase''). Denoting the probability for a change of phase from state $S_0$ to state $S_1$ as $p$, we can characterize the observed reward as
\begin{eqnarray}\label{eq_Reward_random}
r_i\!=\!\left\{\begin{array} {ll}
0, & \textrm{with probability } p,\\
10^{\tilde{T}_i}, & \textrm{with probability } 1-p,
\end{array} \right. 
\end{eqnarray}
where $\tilde{T}_i$ is itself the random variable of the throughput changing due to the change in interference from other SN transmissions when the environment remains in state $S_0$ (there is no change of phase). Examining \eqref{eq_Reward_random}, and considering that interference to the SN link from other SN transmissions is affected by the usually large channel attenuation (in the sense of reducing the transmit power by several orders of magnitude), we posit that the dominant effect determining randomness in the observation of the rewards is the change of phase of the environment from state $S_0$ to state $S_1$ (because it changes the throughput from a value $10^{T_i}$ to a value of zero). As a consequence of this observation we can simplify our model and characterize the observed reward as a Bernoulli random variable given by,
\begin{eqnarray}\label{eq_Reward_random_simpler}
r_i\!=\!\left\{\begin{array} {ll}
0, & \textrm{with probability } p,\\%[-.03in]
K, & \textrm{with probability } 1-p,
\end{array} \right. 
\end{eqnarray}
where $K=\mathbb{E}[10^{\tilde{T}_i}]$ is the expected value of the random variable $10^{\tilde{T}_i}$. 
While in the above modeling the notation for the change of phase probability, $p$, does not indicate dependence on the taken action, in reality this probability indeed depends on the action. To see this, recall that the Q-values are the expected cumulative rewards when starting at a given state and taking first a certain action. Now, consider how the Q-values relate to the actions. Because larger transmit powers result in larger SINR and throughput, if we ignore the action of transmit power equal to zero, the Q-values will be increasing in value for actions associated with larger transmit powers up to the point where the actions are associated with transmit powers so large that they create a change of phase in the environment into state $S_1$ (interference limit on the PN is being violated) and the corresponding Q-values become equal to zero. In other words, the largest Q-value for a CR is associated with its action corresponding with the largest power it can transmit at without exceeding the interference limit (relative throughput change) on its ``nearest'' PN link (the one that is received with largest power). This action with the largest Q-value is the one that should be chosen as the policy at the end of Q-learning. Moreover, relating to \eqref{eq_state_def}, we can write the relative throughput change at the ``nearest'' PN link as $|T_{\%}^{(i)}| = \epsilon + \delta_M(a_t)$, where $\delta_M(a_t)$ is the margin in relative throughput change that is available before a change of phase to state $S_1$ is triggered. This margin depends on the action because larger transmit powers will be associated with smaller margins. Considering now the multi-agent setting, the change of phase in the environment as measured by one CR is primarily determined by its transmit power but is also influenced by the transmissions of other CRs. In one potential scenario, which we have associated above with a probability of occurrence $p$, a CR may be choosing an action (transmit power) for which the environment would stay in state $S_0$ but because of the added effect of other CR transmissions, a change of phase to state $S_1$ occurs. This change of phase effect has a larger probability of occurrence with larger transmit powers resulting in environment state $S_0$ in a single agent case because the relative throughput change margin $\delta_M(a_t)$ is smaller in these cases. Consequently, the probability of change of phase to state $S_1$ increases as the relative throughput change margin $\delta_M(a_t)$ decreases, or equivalently, as the transmit power increases while the environment remains in state $S_0$ if there was a single agent. Moreover, the largest such transmit power, and consequently larger probability $p$, is  associated with the would-be policy after learning. In the following, since our interest is to study convergence to the correct policy, and because this policy is associated with the largest probability to change phase to state $S_1$, we will exclusively consider the probability $p$ as the one associated with the correct policy. 

Examining again \eqref{grad_dec}, we note that because the observed reward $r_i$ is a Bernoulli random variable, the target values used in the DQN training will themselves be random variables. As a result, the Q-value approximations calculated by the DQN will be  random variables themselves. The DQL algorithm will fail to identify the best policy when, after learning, the realization of the Q-value associated with the best policy is less than the Q-value associated with another action. Using Chebyshev's inequality, \cite{ref32}, the probability of this event can be upper bounded in a magnitude proportional to the variance of the best policy's Q-value. Since increasing the variance of the observed reward will increase the variance of the best policy's Q-value, we can equally say that the probability of the best policy's Q-value changing by a magnitude large enough to make it smaller than another Q-value is upper bounded by the variance of the observed reward $\sigma^2_r$. Given that the observed reward has been characterized as a Bernoulli random variable in \eqref{eq_Reward_random_simpler}, its variance is,
\begin{eqnarray}\label{rewardvariance}
\sigma_r^2 = K^2p(1-p).
\end{eqnarray}

\begin{figure}[tb]
	\centering
	\includegraphics[width=0.5\textwidth]{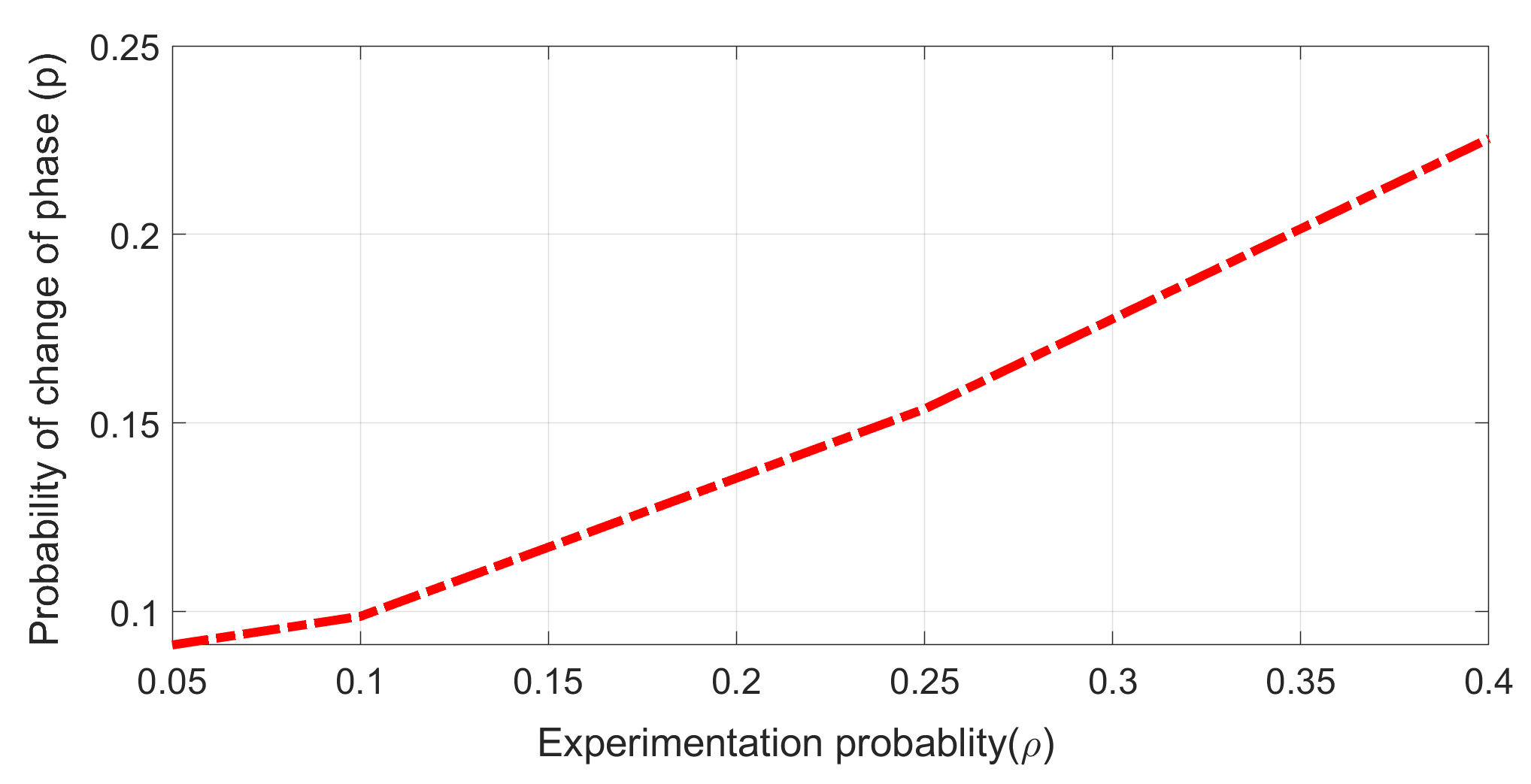}%\vspace*{-1mm}
	\caption{Probability of change of phase $p$ as function of experimentation probability $\rho$.}
	\label{lambda_vs_p}%\vspace*{-1mm}
\end{figure}

This variance could be as small as wanted by having a probability $p$ as small as needed. For a given agent, the value of $p$ is determined by the  margin in relative throughput change $\delta_M(a_t)$ in relation with the level of activity of the other agents. The probability $p$ will increase when the other agents experiment more frequently with actions that are different from their policy in a given exploration phase because more experimentation increases the likelihood of trying actions that will result in a change of phase to state $S_1$. Since the experimentation frequency of different actions is controlled by the experimentation probability $\rho$, the probability $p$ can be made arbitrarily small by choosing an experimentation probability $\rho$ that is sufficiently small. Indeed, this can be readily verified by measuring $p$ as a function of $\rho$ as shown in Fig. \ref{lambda_vs_p}. The figure shows the expected monotonically increasing relation between $p$ and $\rho$. Therefore, we conclude that the probability of having a best policy's Q-value smaller than another action's Q-value can be made arbitrarily small by choosing an exploration probability $\rho$ that is small enough. This guarantees that our distributed and uncoordinated DQL algorithm will approximate the Q-values with sufficient accuracy to converge to the deterministic equilibrium policy. Finally, note also that this reasoning makes implicit use of the fact that of all the Q-values, it is the best policy's Q-value the one with a larger variance because it is associated with the largest probability of a change of phase to state $S_1$ due to the actions of the other agents. 

In conclusion, we have shown that the use of exploration phases allows to control the randomness in the observed reward that originates in the uncoordinated choices of actions of all agents. This control permits the approximation with sufficient accuracy of Q-values using a DQN so that the best policy's Q-value will have the largest of all Q-values. In addition, the exploration phases together with the best reply process with inertia mechanism manage the non-stationarity of the environment to the extent of guaranteeing (with probability one) convergence to the deterministic equilibrium policy.

\section{Experimental Results and Discussion}\label{simulation}%\vspace*{-1mm}

Our focus in the evaluation of the proposed multi-agent DQL technique is twofold. We are first interested in evaluating the convergence to equilibrium states that yield optimal solutions when possible to define such. Secondly, we are interested in the learning performance, which we evaluate as the time to learn the solution policy compared to the table-based approach. All evaluation experiments are based on Monte Carlo simulations. Each simulation instantiates an scenario that comprises of a PN that shares through underlay DSA a 180 kHz radio spectrum band with a SN formed by CRs. The PN consists of nine access points (APs) organized in a three-by-three grid (wrapping around all edges to avoid edge effects), with each AP separated by a distance of 200 m from its neighbors. Of the nine APs, only seven (chosen at random for each Monte Carlo run) are active, each transmitting to one receiver located at random within the coverage area of the corresponding AP. The SN consists of two point-to-point links with each of the two CR transmitters located randomly within the PN grid and the two receivers positioned anywhere within 50 m of their corresponding transmitters. This setup is intended to broadly model a network of small radio devices (the SN) sharing spectrum with an incumbent network (the PN) of larger devices. For underlay spectrum sharing, the limit on relative throughput change on the PN links is set to 5\%.

\begin{figure}[tb]
	\centering
	\includegraphics[width=0.48\textwidth]{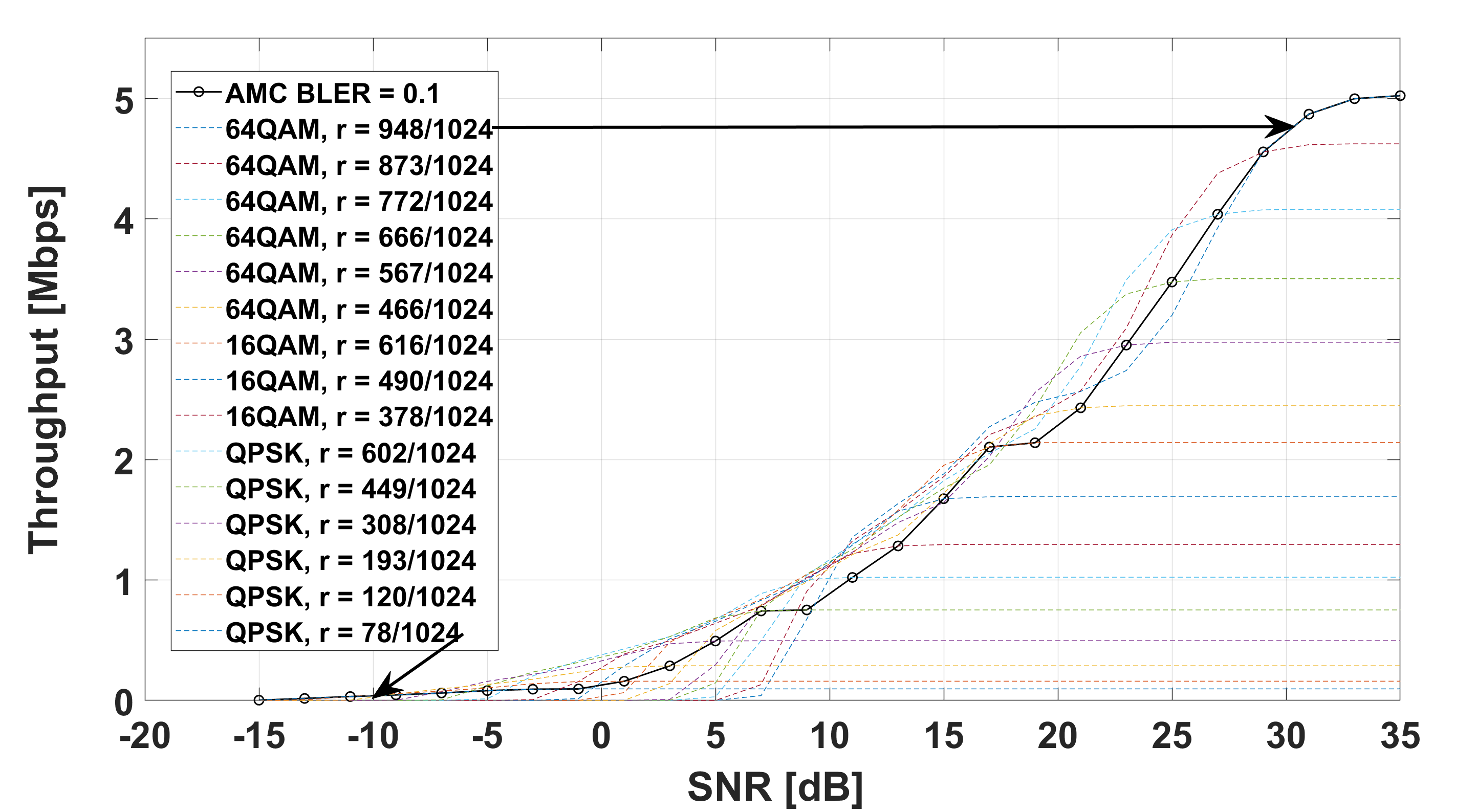}%\vspace*{-1mm}
	\caption{Throughput for each CQI of an LTE system under a Pedestrian B channel.}
	\label{PedB}
\end{figure}
In both networks, the transmitted signals undergo path, penetration and shadowing loss, following the model in \cite{ref33} where the received signal power $P_{Rx}$, the transmitted signal power $P_{Tx}$, the transmitter-receiver distance in km $d$, and the shadowing loss $S$ are related as $P_{Rx} = P_{Tx} - 128.1 -37.6 \log d -10 -S$. In this model, the shadowing loss is modeled as a zero-mean Gaussian random variable with 6 dB standard deviation, the penetration loss is fixed at 10 dB and $P_{Rx}$ and $P_{Tx}$ are measured in dBs. The AWGN power for all the links is set to -130 dBm. As previously said, all transmissions make use of AMC. In the simulation we adopted the LTE AMC scheme (consisting of fifteen possible modes made from the combination of different channel coding rates with one of QPSK, 16-QAM or 64-QAM modulation schemes), \cite{ref34}. This AMC scheme was implemented through the throughput versus SNR AMC performance curves that in our case were obtained using the MATLAB LTE Link Level Simulator from TU-Wien \cite{ref35} and are shown in Fig. \ref{PedB}. These curves, and our simulation model too, include the Pedestrian B small scale fading channel model from \cite{ref36}. Furthermore, we assume that the PN transmitters can adapt transmit power in the range of -20 to 40 dBm following the iterative power allocation algorithm in \cite{ref37} and the CRs operate with an action space, defined as the fourteen transmit power levels between -10 and 20 dBm equally spaced by 2.5 dBm, plus no transmission (transmit power zero). 

Following the evaluation of multiple configurations, we implement the DQN as a four-layer multi layer perceptron (MLP) with input being the environment state and output being the Q-values for each possible action (i.e. the DQN has fourteen outputs). The MLP has eight and eighteen neurons in the hidden layers (from input to output) and the activation function for all hidden layers is saturated ReLU (rectified linear unit). The output layer has a linear activation function. Before delving into the detailed evaluations, our first results explore the effect of exploration phases in controlling the variance of Q-values, as was discussed in Section \ref{sect:convergence}. Our interest here is to see how the variance of Q-values could lead to learning results where the Q-value that is estimated with the largest value is not the one associated with the best action, and then see how the use of exploration phases could be used to avoid this error. To see this, then, we modified the reward function in such a way that, while maintaining the characteristics of our design, it resulted in a global optimal solution. The global optimal reward was defined as
\begin{eqnarray}\label{eq_Reward_global}
R^{(i)}_t(a_t, s_t)\!=\!\left\{\begin{array} {ll}
10^{\sum_{j=1}^N T_j}, & \textrm{if environment state is } S_0,\\%[-.03in]
0, & \textrm{if environment state is } S_1,
\end{array} \right. %\\ [-.13in] \nonumber
\end{eqnarray}
Note that this reward function assumes that CRs can inform of their throughput to all other CRs, but we emphasize that this departure from the uncoordinated scenario is solely to instantiate the particular solution of a global optimum without affecting the nature of our algorithm. Moreover, to focus on the effect of using exploration phases, we consider a simplified version of our DQL algorithm with fix learning rate at 0.01 and we compare its performance with and without using exploration phases (the case without exploration phases is equivalent to doing exploration phases of length $L_E=1$-see Algorithm \ref{alg1}). Other settings in this experiment were number of exploration phases equal to 60, mini-batch size of 25, experimentation probability $\rho = 0.15$, inertia probability $\lambda = 0.3$, and update target Q-values every c = 30 training steps.

\begin{figure}[tb]
	\centering
	\includegraphics[width=0.5\textwidth]{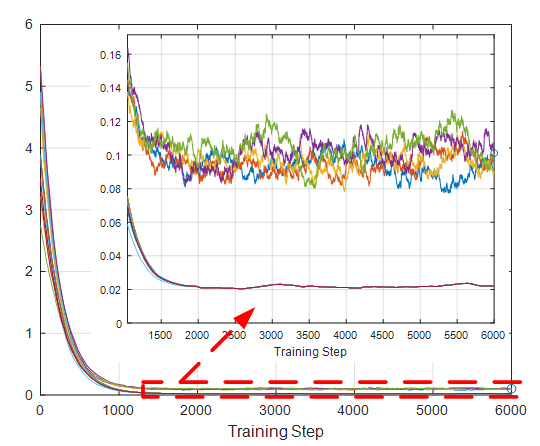}%\vspace*{-1mm}
	\caption{Q-values evolution during multi-agent learning without exploration phases.}
	\label{singleagent}%\vspace*{-1mm}
\end{figure}

\begin{figure}[tb]
	\centering
	\includegraphics[width=0.5\textwidth]{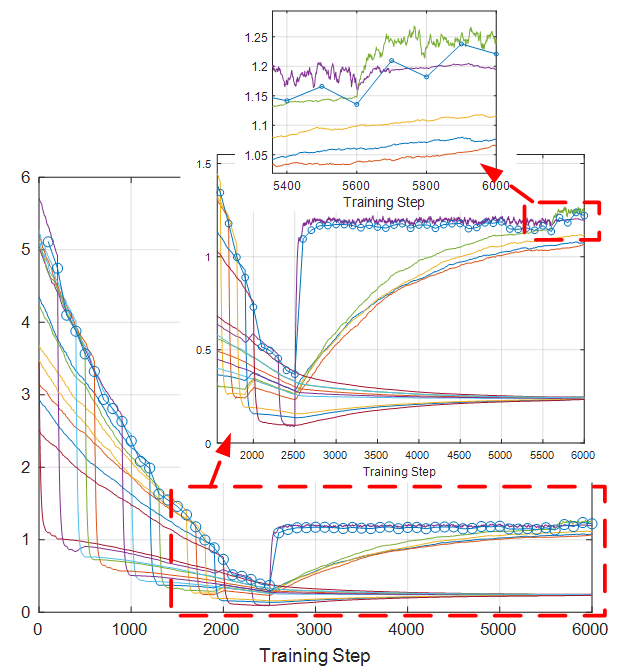}%\vspace*{-1mm}
	\caption{For identical settings as in Fig. \ref{singleagent}, Q-values evolution using the proposed DQL approach with exploration phases (solution converges to optimal policy).}
	\label{multiagent}%\vspace*{-1mm}
\end{figure}
Figs. \ref{singleagent} and \ref{multiagent} compare the evolution of Q-values during learning for the DQL approach without and with exploration phases, respectively. Both figures show RL evolution for CR link 2, the evolution over the training steps of all the Q-values when in state $S_0$ and for the same exemplar realization of the system (same network setup, channel gains, initialization, etc.). The figures include a curve with circular markers, which shows the evolution of $\max_{v^i}Q_{t_{k+1}}^i (s, v^i)-\delta^i$ in line \ref{startinert} of Algorithm \ref{alg1}. We call this curve as the ``candidate policy set'' curve because action associated with a Q-value larger than this curve could be chosen as the next policy. Inserts in the figures show zoomed-in portions of the curves. For DQL without exploration phases, Fig. \ref{singleagent} shows how the variance in the Q-values results in their values for multiple actions being indistinguishable from each other and leading to the impossibility of convergence to a solution (or a choice for the final policy that becomes practically a random choice between different actions). In addition, it is important to note that the configuration depicted in Fig. \ref{singleagent} makes use of the customary separate target action-value updated every $c=60$ training steps. As we discussed in Sect. \ref{clcr}, this feature which addresses the non-stationarity of the action-value function could on the surface have been thought to be able to address other non-stationary elements, in particular the environment. However, Fig. \ref{singleagent} clearly shows that this is not the case and that DQL without exploration phases may not achieve convergence in an uncoordinated multi-agent scenario.

In contrast, Fig. \ref{multiagent} depicts the learning evolution under completely identical settings for our proposed multi-agent DQL technique with exploration phases and illustrates success in achieving convergence to the optimal solution. As can be seen in the zoomed-in details of the curve (two zoom levels are shown in the figure), the use of exploration phases achieves a near-stationary environment that results in the reduction in the variance of Q-values for as much as is needed to obtain discernible and accurate estimates. Now, the best action has a Q-value estimate that is correctly the largest and, while still showing some ``noise'', it does not ``overlap'' with other Q-values curves. This behavior, of course, leads to a proper convergence to the optimal solution. It is also of interest to note in Fig. \ref{multiagent} that the Q-value for the policy has a larger variance than the Q-values for other actions. This observation is a confirmation of two arguments used in the study presented in Sect. \ref{sect:convergence} because it is due to the fact that the best policy has the smallest relative throughput change margin $\delta_M(a_t)$ and is also subject to the effects of the other CRs more frequently (because of being the policy during an exploration phase). 

Lastly with regards to Figs. \ref{singleagent} and \ref{multiagent}, it is interesting to note in Fig. \ref{singleagent} the inaccurate calculation of many of the Q-values, with results much smaller (near zero) to what they should be (as shown in Fig. \ref{multiagent}). This is because the scenario depicted in the figures is one where the other CR frequently tests actions that triggers for CR link 2 an environment phase transition to state $S_1$ (large probability $p$ for change of phase to state $S_1$). This results in CR link 2 observing for an action that should have a non-zero reward, a value of zero in the majority of cases. Fig. \ref{multiagent} shows that by controlling the exploration probability ($\rho=0.15$ in this case) the probability $p$ for change of phase to state $S_1$ is reduced to the level that most of the time the system is kept in state $S_0$ and yields a reward larger than zero (at it should be). 

\begin{figure}[tb]
	\centering
	\includegraphics[width=0.5\textwidth]{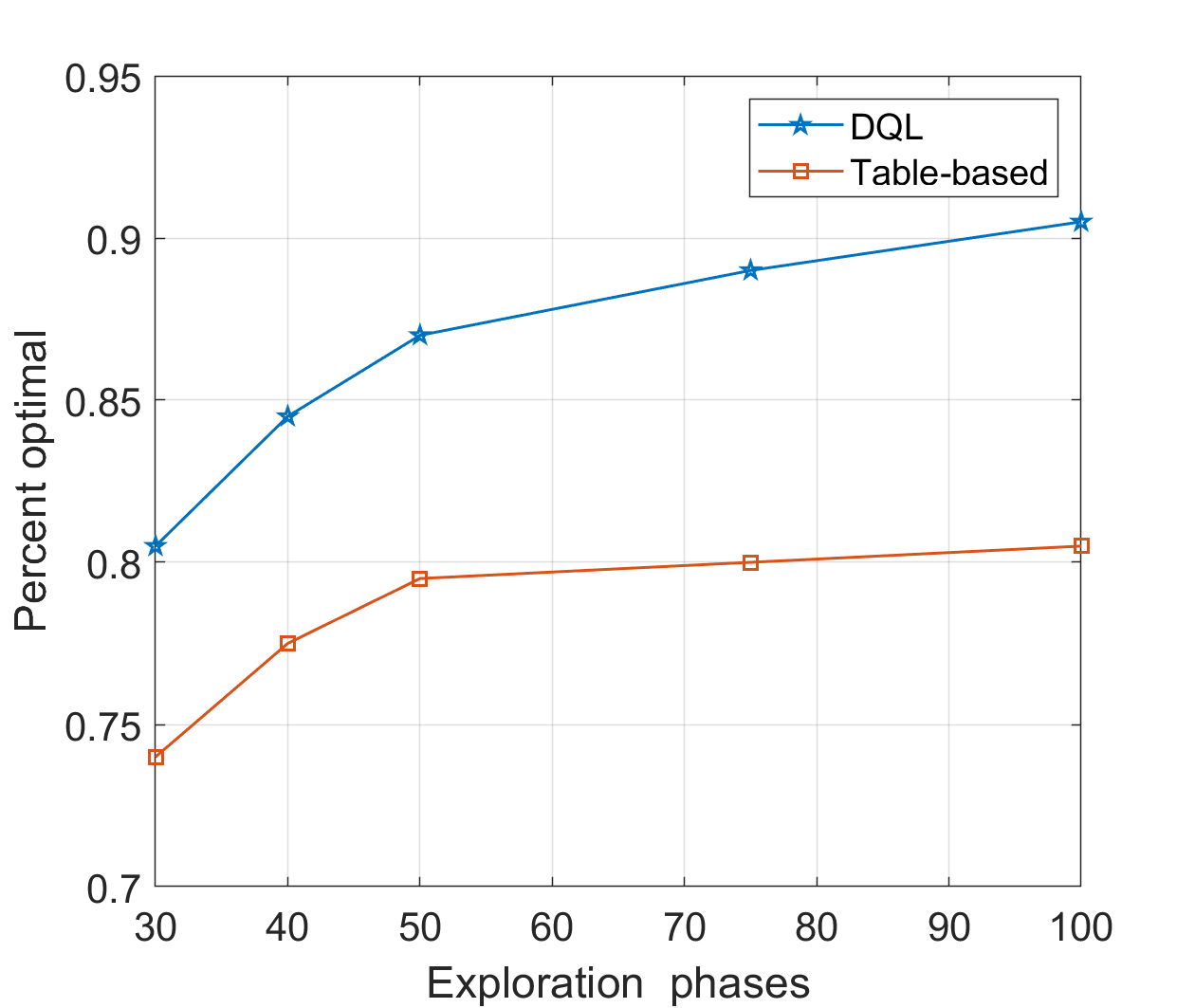}%\vspace*{-0.2mm}
	\caption{Learning performance comparison between table-based approach and proposed DQL algorithm.} %\vspace*{-1mm}
	\label{comparison}
\end{figure}

We now proceed to study the learning performance of our proposed DQL technique and to observe its convergence to equilibrium states. In order to unambiguously be able to identify through exhaustive search the converged state as an equilibrium one, in this study we continue to use the reward \eqref{eq_Reward_global} that makes the equilibrium state be the global optimum solution. We evaluate the learning performance by measuring the percentage of times that the DQL process finds the optimal solution after running for a number of exploration phases. We first focus the study on the learning performance compared to the equivalent table-based realization. Figure \ref{comparison} shows the result of this aspect of the performance study by comparing the curves ``DQL'' (our DQL algorithm) and ``Table-based'' (the equivalent table-based realization). Each point in either curve is the percentage of times over 100 Monte Carlo runs that the RL algorithm found the optimum solution after learning for a number of exploration phases. The optimum solution was identified by performing an exhaustive search over all possible combination of actions for all CRs ($|\mathcal{A}|^N=196$ combinations in our case) for the network setup in each Monte Carlo run.

To have a fair comparison, the equivalent table based performance curve was realized for the same number of learning steps as for the DQL curve. One exploration phase in the DQL result consists of 250 mini-batch learning instances with mini-batch size of 25, resulting in an exploration phase length of 6250 learning steps (taking a single action and receiving the corresponding reward). Consequently, for fair comparison the exploration phase length in the table-based case was set directly to 6250 learning steps because the learning algorithm in this case does not uses mini-batches of training samples. Besides the common fixed setting of exploration phases length, for other hyper-parameters we explored multiple combinations of choices and we selected the ones yielding highest percentage of optimal solutions for each point. Table \ref{DQLparamnoann} shows the selected hyper parameters for our DQL algorithm. For the table-based case we selected for exploration phases $\alpha=0.2$, $\zeta=3$, $\rho=0.2$, and $\lambda=0.25$.

\begin{figure}[tb]
	\centering
	\includegraphics[width=0.5\textwidth]{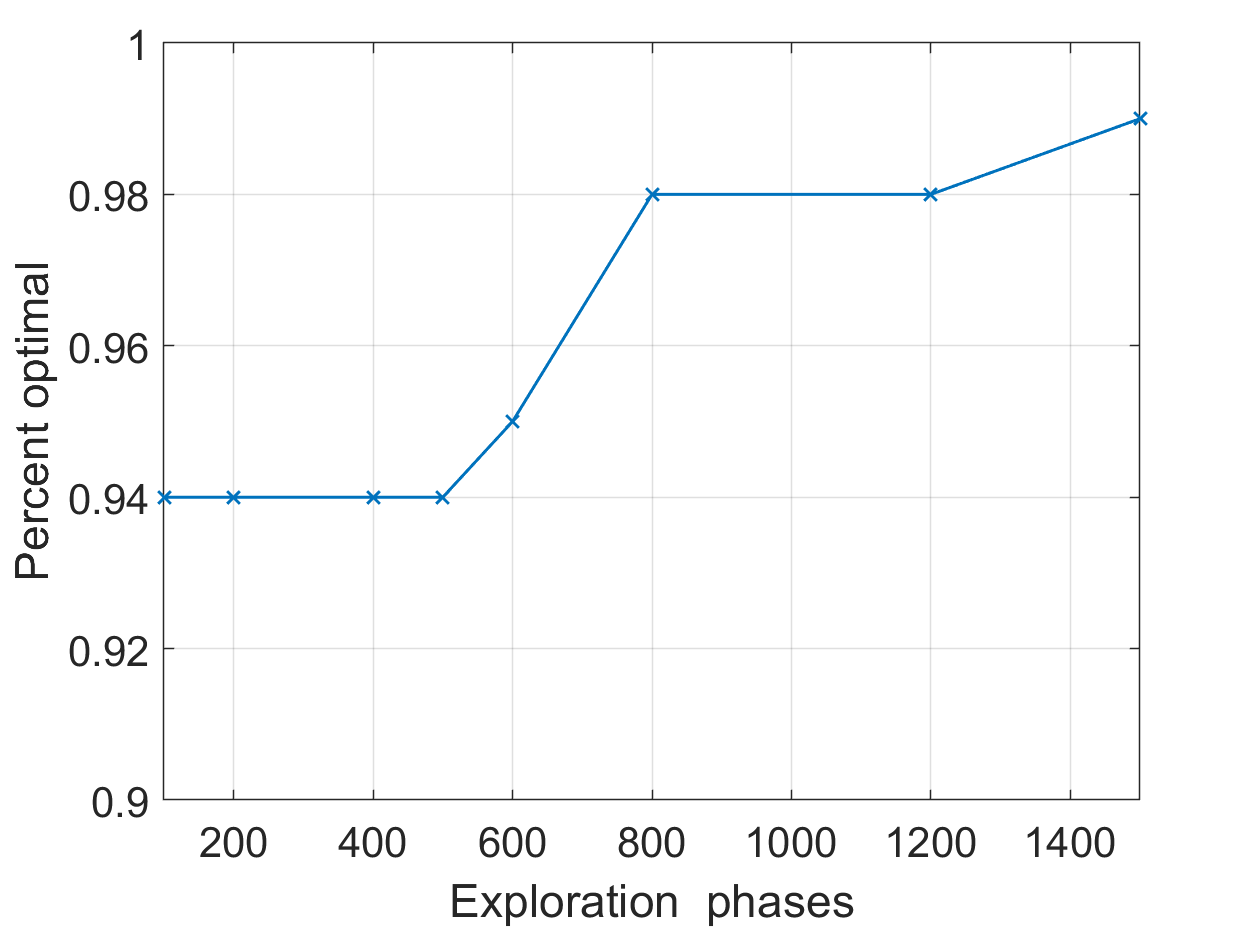}
	\caption{Convergence to optimal policy as a function of exploration phases.} 
	\label{ver_optimal}
\end{figure}

Figure \ref{comparison} shows that our proposed DQL algorithm learns significantly faster than the equivalent table-based approach. Equivalently, Fig. \ref{comparison} shows that compared to the table-based approach, our DQL algorithm achieves a higher percentage of optimal solutions when learning is stopped after a certain number of exploration phases.
\begin{table}[tb]
	\centering
	\caption{Hyper-parameters for points of curve ``DQL'' in Fig. \ref{comparison}.} \label{DQLparamnoann}	
	\begin{tabular} { | c | c | c | c | c | c |  }
		%{c c }
		\hline 
		\thead{Number of \\ Exploration \\ Phases} & \thead{ \\ $\alpha$ \\ } & \thead{ \\ $\zeta$ \\ } & \thead{ \\ $\rho$ \\ } & \thead{ \\ $\lambda$ \\ } & \thead{ \\ $c$ \\ } \\  %[-0.6ex]
		\hline		
		30 & 0.050 & 5.0 & 0.10 & 0.25 & 50  \\
		\hline
		40 & 0.030 & 5.0 & 0.20 & 0.25 & 50  \\
		\hline
		50 & 0.050 & 4.5 & 0.20 & 0.25 & 50  \\
		\hline
		75 & 0.015 & 2.0 & 0.05 & 0.25 & 50  \\
		\hline
		100 & 0.050 & 5.0 & 0.10 & 0.25 & 50  \\
		\hline
	\end{tabular}%\vspace{1mm}
\end{table}
After comparing the learning performance of our proposed DQL algorithm to the equivalent table-based approach we now switch focus to empirically verify the convergence of our proposed DQL algorithm to an equilibrium state (now realized as a global optimum state still identified through exhaustive search). We achieve this by measuring the percentage of Monte Carlo runs that our DQL algorithm finds the optimum solution after learning for a number of exploration phases. In the process of this experiment we identified a rare phenomenon (we measured it to occur in around 4\% of cases) related to the particular characteristics of how throughput changes as a function of the link SINR (as illustrated in Fig. \ref{PedB}). The phenomenon occurs when both CRs links need to operate at low SINR and there is a ``symmetry'' in the optimal solution between the two CR links. To see an example of a potential such case consider that the fourteen possible actions are sorted in order of increasing transmit power (no transmission is action one, transmit power of -10 dBm is action 2, transmit power of -7.5 dBm is action 3, etc.) In the example, the optimal solution is for both CR links operating at low SINR with optimal transmit power for CR 1 being action 1 and optimal transmit power for CR 2 being action 2. In a network scenario as this one, if CR 1 takes action 2 and CR 2 takes action 1 (a swap of actions from the optimum between the two CRs), the resulting reward \eqref{eq_Reward_global} is very close to the optimal reward (we observed relative differences usually orders of magnitude smaller than 1\%). This very small difference in rewards may lead for certain initialization of the learning to the algorithm learning as policy the actions associated with the near-optimum result. This situation could be avoided by reducing the exploration probability to very small values. However, this solution does not appear as a sensible approach given that it is required by a small percentage of Monte Carlo runs (4\% or less) and that the reduction of the exploration probability to the small values that would be needed leads to a large increase in learning time (in essence, a confirmation that a very small percentage of setups require very long convergence times). Instead, we resolve these cases by a simple add-on to our algorithm which in some cases increases the learning time by negligible amounts (e.g. less than 3\% increase in the case when reaching 99\% of optimal solutions). In this add-on, the algorithm repeats four times a short learning process (for just ten exploration phases) that each starts from a new random initialization and saves the learning state at the end of each initial learning case. After the four short learning processes, the add-on selects the case that has achieved the largest reward and continues the learning process from the state at the end of the corresponding ten exploration phases. With this add-on in place, it can be seen in Fig \ref{ver_optimal} how after 1500 exploration phases our algorithm has learned the optimal policy in 99\% of the cases. In addition, by comparing Fig. \ref{ver_optimal} with Fig. \ref{comparison}, it can be seen how the random re-initialization add-on resolves the cases described above occurring 3.5\% of times.

\vspace*{-2mm}
\section{Conclusion}\label{conclsec}\vspace*{-1mm}

In this paper, we have presented a novel multi-agent deep Q-learning technique that converges to an equilibrium policy (optimal policy when it can be defined) in a distributed and uncoordinated multi-agent wireless network and that is able to learn faster than an equivalent table-based Q-learning implementation. The multi-agent environment is represented by a cognitive radio that coexists with a primary network through underlay dynamic spectrum access. Each CR independently learns the resource (power in this work) allocation decision subject to achieving the largest throughput without violating the relative throughput change threshold at its nearest PN link. The proposed DQL technique succeeds in addressing the challenge of a non-stationary multi-agent environment resulting from the dynamic interaction between radios through the shared wireless environment by combining a deep neural network that uses environment states as the input and outputs Q-values for all the possible discrete actions, with learning in exploration phases, and with the use of a Best Reply Process with Inertia for the gradual learning of the best policy. 
In this work, we first analyze the convergence properties of our multi-agent DQL algorithm addressing two key aspects. The first aspect we consider, common to other approaches of Q-learning, is the policy convergence to equilibrium states in a multi-agent set up. The second aspect, specific to the DQL approach based on approximating a function, studied the effect of noisy reward observations on the function approximation and how the use of exploration phases allows the reduction of the noise variance to arbitrary small values. In this regard, we also show through an example that a DQL realization that does not learn in exploration phases may not be able to converge due to the effect of other agents' actions on the environment seen by an agent of interest. Simulation results show that under sufficiently long learning times the presented technique finds the optimal policy in nearly 99 \% of cases, confirming that our DQL algorithm converges to equilibrium policies for an arbitrarily long learning time. In addition, simulations show that our DQL approach requires less than half the number of learning steps to achieve the same performance as an equivalent table-based implementation.
\vspace*{-1mm}

\begin{IEEEbiographynophoto}{ANKITA TONDWALKAR} received her BE degree in Electronics and Telecommunications Engineering in 2013, and her ME degree in Electronics and Telecommunications Engineering from University Of Mumbai in 2015. She was an ASSISTANT PROFESSOR for three years with the the Electronics and Telecommunications Department, Vidyalankar Institute Of Technology, University of Mumbai, India. Her research interest includes cognitive radios, applications of machine learning to wireless communications.
\end{IEEEbiographynophoto}
\begin{IEEEbiographynophoto}{ANDRES KWASINSKI} (S'98-M'04-SM'11) received  in 1992 his diploma in Electrical Engineering from the Buenos Aires Institute of Technology,Buenos Aires, Argentina, and the MS and PhD degrees in Electrical and Computer  Engineering  from  the  University  of Maryland,  College Park,  Maryland,  in  2000  and 2004, respectively. He is currently a Professor at the Department of Computer Engineering, Rochester Institute  of  Technology,  Rochester,  NY. Dr. Kwasinski has co-authored more than one hundred publications in peer-reviewed journals and international conferences.  He  has  also  co-authored  the  books ``Cooperative  Communications  and  Networking'' (Cambridge  University Press,  2009)  and  ``3D  Visual Communications'' (Wiley, 2013). He is
currently an Associate Editor for the IEEE Signal Processing Magazine and
Chief Editor of the IEEE Signal Processing Society Resource Center. Dr.
Kwasinski has been an Editor for the IEEE Transactions on Wireless Communications and the IEEE Wireless Communications Letters.  His current  areas  of  research  include  cognitive  radios and  wireless networks,  cross-layer  techniques  in  wireless  communications and smart infrastructures and networking.
\end{IEEEbiographynophoto}

\end{document}